%% file: neurips_2021.tex
\documentclass{article}

\usepackage[nonatbib,preprint]{neurips_2021}
\usepackage[numbers]{natbib}

\usepackage[utf8]{inputenc} % allow utf-8 input
\usepackage[T1]{fontenc}    % use 8-bit T1 fonts
\usepackage{hyperref}       % hyperlinks
\usepackage{url}            % simple URL typesetting
\usepackage{booktabs}       % professional-quality tables
\usepackage{amsfonts}       % blackboard math symbols
\usepackage{nicefrac}       % compact symbols for 1/2, etc.
\usepackage{microtype}      % microtypography
\usepackage{xcolor}         % colors
\usepackage{subfig}

\usepackage[british]{babel}

\usepackage{natbib} % has a nice set of citation styles and commands
\usepackage{mathtools} % amsmath with fixes and additions
\usepackage{booktabs} % commands to create good-looking tables
\usepackage{tikz} % nice language for creating drawings and diagrams

\usepackage[utf8]{inputenc}
\usepackage{amsfonts}
\usepackage{xfrac}
\usepackage[colorinlistoftodos]{todonotes}

\usepackage{comment}
\usepackage{amsthm}
\usepackage{subfig}
\usepackage{hyperref}
\usepackage{booktabs}
\usepackage{float}
\usepackage{wrapfig}
\usepackage{empheq}
\usepackage[theorems,skins]{tcolorbox}
\newcommand*\widefbox[1]{\fbox{\hspace{1em}#1\hspace{1em}}}
\newcommand{\inv}{^{\raisebox{.2ex}{$\scriptscriptstyle-1$}}}

\newcommand{\ffrac}[2]{\ensuremath{\frac{\displaystyle #1}{\displaystyle #2}}}

\usepackage[numbers]{natbib}

\newtheorem{theorem}{Theorem}[section]

\newtheorem{proposition}[theorem]{Proposition}

\newtheorem{definition}{Definition}[section]

\usepackage{graphicx}

\title{Hypothesis Testing for Class-Conditional Label Noise}

\author{%
  Rafael Poyiadzi\\
  Department of Engineering Mathematics\\
  University of Bristol\\
  Bristol, United Kindgom \\
  \texttt{rp13102@bristol.ac.uk} \\
   \And
   Weisong Yang \\
  Department of Engineering Mathematics\\
  University of Bristol\\
  Bristol, United Kindgom \\
  \texttt{ws.yang@bristol.ac.uk} \\
   \AND
   Niall Twomey\\
  Cookpad Ltd\\
  \& University of Bristol\thanks{Honorary}\\
  Bristol, United Kindgom \\
  \texttt{niall-twomey@cookpad.com} \\
   \And
   Raul Santos-Rodriguez\\
  Department of Engineering Mathematics\\
  University of Bristol\\
  Bristol, United Kindgom \\
  \texttt{enrsr@bristol.ac.uk} \\
}

\begin{document}

\maketitle

\input{sections/abstract}
\input{sections/introduction}
\input{sections/background}
\input{sections/tests}
% \input{sections/notes}
\input{sections/related_work}

\input{sections/experiments}
\input{sections/conclusion}

\clearpage

\bibliography{bibliography}

\clearpage

\input{sections/appendix}

% \bibliography{uai2021-template}

\clearpage
\end{document}

%% file: sections/abstract.tex
\begin{abstract}
In this paper we provide machine learning practitioners with tools to answer the question: \textit{is there class-conditional noise in my labels?} In particular, we present hypothesis tests to check whether a given dataset of instance-label pairs has been corrupted with \textit{class-conditional label noise}, as opposed to \textit{uniform label noise}, with the former biasing learning, while the latter -- under mild conditions -- does not. The outcome of these tests can then be used in conjunction with other information to assess further steps. While previous works explore the direct estimation of the noise rates, this is known to be hard in practice and does not offer a real understanding of how trustworthy the estimates are. These methods typically require \textit{anchor points} -- examples whose true posterior is either $0$ or $1$. Differently, in this paper we assume we have access to a set of anchor points whose true posterior is approximately $1/2$. The proposed hypothesis tests are built upon the asymptotic properties of Maximum Likelihood Estimators for Logistic Regression models. We establish the main properties of the tests, including a theoretical and empirical analysis of the dependence of the power on the test on the training sample size, the number of anchor points, the difference of the noise rates and the use of relaxed anchors.

\end{abstract}

% \begin{abstract}
% In this work we study the question: \textit{Is there class-conditional flipping noise in my dataset?} Specifically, provided with a dataset $\mathcal{D}_n = \{(\boldsymbol{x}_i, y_i)\}_{i=1}^n$, we want to assess whether the labels we observe are drawn from their true posterior distribution: $Y~|~(X=x) \sim \eta(\boldsymbol{x}) = \mathbb{P}(Y~|~X=x)$, or from a corrupted version of it: $\tilde{Y}~|~(X=x) \sim \tilde{\eta}(\boldsymbol{x}) = (1-\alpha-\beta)\cdot\eta(\boldsymbol{x}) + \beta$, where $\alpha,~\beta$ denote the class-conditional noise rates. As we reference in the follow-up, in the case of \textit{uniform noise} ($\alpha = \beta$), under mild conditions, risk minimisation is robust.

% Many works exist that aim to estimate the noise rates, but there is no understanding of how trustworthy these estimates are. An object of much interest are \textit{anchor-points} - points that are \textit{`perfect'}, i.e. $\eta(\boldsymbol{x}) \in \{0,1\}$. In this work We assume we are provided with a set of \textit{anchor-points} whose true posterior is $\approx \sfrac{1}{2}$. We combine this with asymptotic properties of Maximum Likelihood Estimators to device hypothesis tests to assess whether $\alpha = \beta$. We analyse the dependence of the power of the test on the training sample size, the number of anchor-points, the factors ($\beta-\alpha$) and ($\eta(\boldsymbol{x}) - \sfrac{1}{2}$) and then examine experimentally the claims.
% \end{abstract}

%% file: sections/introduction.tex
\section{Introduction}
\label{section:introduction}

When a machine learning practitioner is presented with a new dataset, a first question is that of data quality (\cite{lawrence2017data}) as this will affect any subsequent tasks and inferences. This has led to tools to address transparency and accountability of data (\cite{gebru2018data, kacper2019fat}). However, in supervised learning, an equally important concern is the quality of labels. For instance, in standard data collections, data curators usually rely on annotators from online platforms, where individual annotators cannot be unconditionally trusted as they have been shown to perform inconsistently (\cite{jindal2017learning}). Labels are also expected to not be ideal in situations where the data is harvested directly from the web (\cite{1544937}, \cite{harvesting_images}). In general this is a product of annotations not being carried out by domain experts.

The existing literature focuses on estimating the distortion(s) present in the labels (see Section~\ref{section:related_work}). In this paper we take a step back and our main contribution is the design and analysis of hypothesis testing procedures that would allow us (under certain assumptions we state later) to provide the practitioner with a measure of evidence for class-conditional noise, against uniform noise (as we discuss later, class-conditional noise biases the learning procedure, while uniform noise under mild conditions  does not). With this information at hand, the practitioner can then make more informed decisions. What we present is designed to be performed after data collection and annotation to offer a quality measure with respect to label noise. If the quality is deemed poor, then the practitioner could resort to: (1) a modified data labelling procedure (e.g., active learning in the presence of noise), (2) seek methods to make the training robust (e.g., algorithms for learning from noisy labels), or (3) drop the dataset altogether. 

In binary classification, the goal is to train a classifier $g: \mathcal{X} \to \{-1,+1\}$, from a labelled dataset $\mathcal{D}^{train}_n = \{(\boldsymbol{x}_i, y_i)\}_{i=1}^n  \in (\mathbb{R}^d\times \{-1, 1\})$, with the objective of achieving a low miss-classification error: $\mathbb{P}_{X, Y}(g(X) \neq Y)$. While it is generally assumed that the training dataset is drawn from the distribution for which we wish to minimise the error for $\mathcal{D}^{train}_n \sim p(X, Y)$, this is often not the case. Instead, the task requires us to train a classifier on a corrupted version of the dataset $\tilde{\mathcal{D}}^{train}_n \sim p(X, \tilde{Y})$ whilst still hoping to achieve a low error rate on the clean distribution. In this work we focus on one particular type of corruption: \textit{instance-independent label noise}, where labels are flipped with a certain rate, that can either be uniform across the entire data-generating distribution or conditioned on the true class of the data point. A motivating example of class-conditional noise is given in \cite{frenay2013classification} in the form of medical case-control studies, where different tests may be used for subject and control.

An essential ingredient in our procedure is the input from the user in the form of a set of \textit{anchor points}. Differently from previous works, we assume anchor points for which the true posterior distribution $\mathbb{P}(Y=1~|~X=x)$ is (approximately) \sfrac{1}{2}. For an instance $\boldsymbol{x}$ this requirement means that an expert would not be able to provide \textit{any} help to identify the correct class label. While this will be shown to be convenient for theoretical purposes, finding such anchor points might be rather difficult to accomplish in practice, so we show how to relax this notion to a more realistic $\eta(x) \approx \sfrac{1}{2}$. Anchor-points need to be provided by the experts.

Our approach is based on the asymptotic properties of the \textit{Maximum Likelihood Estimate} (MLE) solution for Logistic Regression models, and the relationship between the true and noisy posteriors. On the theoretical side, we show that when the asymptotic properties of MLE hold and the user provides a single anchor point, we can devise hypothesis tests to assess the presence of class-conditional label corruption in the dataset. We then further extend these ideas to allow for richer sets of anchor points and illustrate how these lead to gains in the \textit{power} of the test. 

% Defining the \textit{null hypothesis} as having uniform noise (or, no noise), we show how to utilise anchor-points to obtain a continuous measure of evidence against the null: the \textit{p-value} - which is more aligned with Fisher's significance testing. But, we also define the implicit \textit{alternative hypothesis} of class-conditional noise, and proceed to introduce a \textit{significance level} and analyse the \textit{power} of the test - which is more aligned with Neyman-Pearson theory, where the p-value is the basis of formal decision making process of rejecting, or failing to reject, the null. Both the p-value and the output of the test can be used as part of a broader decision process that considers other important factors.

In Section \ref{section:background} we cover the necessary background on MLE, noisy labels and define the necessary tools. In Section \ref{section:hypothesis_tests} we illustrate how to carry a $z$-test using anchor points on the presence of class-conditional noise. In Section \ref{section:related_work} we discuss related work and in Section \ref{section:experiments} we present experimental findings.

%% file: sections/background.tex
\section{Background}
\label{section:background}

We are provided with a dataset $(\boldsymbol{X}, \boldsymbol{y}) = \{(\boldsymbol{x}_i, y_i)\}_{i=1}^N \in (\mathbb{R}^d\times \{-1, 1\})$, and our task is to assess whether the labels have been corrupted with class-conditional flipping noise. We use $y$ to denote the true label, and $\tilde{y}$ to denote the noisy label. We assume the feature vectors ($\boldsymbol{x}$) have been augmented with $ones$ such that we have $\boldsymbol{x} \to (1,~\boldsymbol{x})$. We assume the following model:
\begin{equation*}
y_{i} \sim \text {Bernoulli}\left(\eta_{i}\right),~~~~\eta_{i}=\sigma(\theta_0^\top \boldsymbol{x}_{i})=\frac{1}{1+\exp \left(-\theta_0^\top \boldsymbol{x}_{i}\right)}
\end{equation*}

Following the MLE procedure we have: $\hat{\theta}_{n} := \underset{\theta \in \Theta}{\operatorname{argmax}}~\ell_n\left(\theta \mid D_n\right) = \underset{\theta \in \Theta}{\operatorname{argmax}}~\prod_{i=1}^n~\ell_i\left(\theta \mid \boldsymbol{x}_i,~y_i\right)$

% \begin{equation*}
% \hat{\theta}_{n} := \underset{\theta \in \Theta}{\operatorname{argmax}}~\ell_n\left(\theta \mid D_n\right) = \underset{\theta \in \Theta}{\operatorname{argmax}}~\prod_{i=1}^n~\ell_i\left(\theta \mid \boldsymbol{x}_i,~y_i\right)
% \end{equation*}
where: $l\left(\theta \mid \boldsymbol{x}_i,~y_i\right) = \frac{y_i+1}{2}\cdot\log\eta_i + \frac{1-y_i}{2}\cdot\log(1-\eta_i)$. In this setting, the following can be shown (See for example Chapter 4 of \cite{van2000asymptotic}):
\begin{equation}
\label{eq:mle_asymptotic}
\sqrt{n}\left(\hat{\theta}_{n}-\theta_{0}\right)~\stackrel{D}{\longrightarrow} \mathcal{N}\left(0,~I_{n}(\theta_0)\inv\right)
\end{equation}

where $I_{\theta_{0}}$ denotes the Fisher-Information Matrix:
\begin{equation*}
I_{n}(\theta_0)=\mathbb{E}_{\theta}\left(-\frac{\partial^{2} \ell_{n}(\theta ; Y \mid x)}{\partial \theta \partial \theta^{\top}}\right)=\mathbb{E}_{\theta}\left(-H_{n}(\theta ; Y \mid x)\right)
\end{equation*}
where the expectation is with respect to the conditional distribution, and $H_n$ is the Hessian matrix. 

% An important remark is that we can express $I_{n}(\theta_0) = n\cdot I_{i}(\theta_0)$, which shows that as $n \to \infty$, the variance of the MLE estimator goes to $0$.

We will consider two types of flipping noise and in both cases the noise rates are independent of the instance: $\mathbb{P}(\tilde{Y} = -i~|~Y=i,~X=x) = \mathbb{P}(\tilde{Y} = -i~|~Y=i)$ for $i \in \{-1, 1\}$.

\begin{definition}
\emph{Bounded Uniform Noise (UN)}\\
\label{def:uniform_noise}
In this setting the per-class noise rates are identical: $\mathbb{P}(\tilde{Y} = 1~|~Y=-1) = \mathbb{P}(\tilde{Y} = -1~|~Y=1) = \tau$ and bounded: $\tau < 0.50$. We will denote this setting with UN($\tau$), and a dataset $\mathcal{D} = (\boldsymbol{X}, \boldsymbol{y})$ inflicted by UN($\tau$) by: $\mathcal{D}_{\tau}$. 
\end{definition}

\begin{definition}
\emph{Bounded Class-Conditional Noise (CCN)}\\
\label{def:classconditional_noise}
In this setting the per-class noise rates are different, $\alpha \neq \beta$ and bounded $\alpha + \beta < 1$ with: $\mathbb{P}(\tilde{Y} = -1~|~Y=1) = \alpha$ and $\mathbb{P}(\tilde{Y} = 1~|~Y=-1) = \beta$. We will denote this setting with  CCN($\alpha, \beta$), and a dataset $\mathcal{D} = (\boldsymbol{X}, \boldsymbol{y})$ inflicted by CCN($\alpha, \beta$) by: $\mathcal{D}_{\alpha, \beta}$.
\end{definition}

% \begin{figure}
%     \centering
%     \includegraphics[width=0.90\textwidth]{figures/noise_transform.pdf}
%     \caption{An illustration of how CCN alters the class distributions of two Gaussian class-conditional distributions.}
% \end{figure}
 
% \begin{figure}
%     \centering
%     \includegraphics[width=0.90\textwidth]{figures/decision_boundary.pdf}
%     \caption{\textit{Top Left:} Original (clean) data, and \textit{Top Right:} we have the noisy version. At the bottom we have the two corresponding learned decision boundaries, with the crosses corresponding to a set of anchor points.}
% \end{figure}

An object of central interest in classification settings is the posterior predictive distribution: $\eta(\boldsymbol{x}) = \mathbb{P}(Y = 1~|~X=\boldsymbol{x})$. Its noisy counterpart, $\tilde{\eta}(\boldsymbol{x})~=~\mathbb{P}(\tilde{Y} = 1~|~X=\boldsymbol{x})$, under the two settings, $UN(\tau)$ and $CCN(\alpha, \beta)$, can be expressed as: (See Appendix 8.1 for full derivation)
\begin{equation}
    \label{eq:noisy_posterior}
    \tilde{\eta}(\boldsymbol{x})~=\left\{\begin{array}{ll}
(1-\alpha-\beta)\cdot\eta(\boldsymbol{x}) + \beta & \text { if } \textrm{(CCN)} \\
(1-2\tau)\cdot\eta(\boldsymbol{x}) + \tau & \text { if } \textrm{(UN)} \\
\end{array}\right.
\end{equation}

We consider loss functions that have the margin property: $\ell(y, f(x)) = \psi(yf(x))$, where $f: \mathbb{R}^d\to\mathbb{R}$ is a scorer, and $g(\boldsymbol{x})=sign(f(\boldsymbol{x}))$ is the predictor. Let $f^{*} = \arg\min_{f\in\mathcal{F}}~ \mathbb{E}_{X, Y} \psi(Yf(X))$ and $\tilde{f}^{*} = \arg\min_{f\in\mathcal{F}}~ \mathbb{E}_{X, \tilde{Y}} \psi(\tilde{Y}f(X))$ denote the minimisers under the clean and noisy distributions, under model-class $\mathcal{F}$.

\begin{definition}
\emph{Uniform Noise robustness (\cite{ghosh2015making})}\\
 Empirical risk minimization under loss function $\ell$ is said to be noise-tolerant if
$\mathbb{P}_{X, Y}(g^{*}(X)=Y)=\mathbb{P}_{X, Y}(\tilde{g}^{*}(X)=Y)$.
\end{definition}

\begin{theorem}
\emph{Sufficient conditions for robustness to uniform noise}\\
\label{theorem:un_robustness}
Under uniform noise $\tau < 0.50$, and a margin loss function, $\ell(y, f(x)) = \psi(yf(x))$ satisfying: $\psi(f(X)) + \psi(-f(X)) = K$ for a positive constant $K$, we have that $\tilde{g}^*(x) = sign(\tilde{f}^*(x))$ obtained from: $\tilde{f}^* = \arg\min_{f\in\mathcal{F}}~\mathbb{E}_{X, \tilde{Y}} \psi(\tilde{Y}f(X))$ is robust to uniform noise.
\end{theorem}
For the proof see Appendix 8.2. Several loss functions satisfy this, such as: the \textit{square}, \textit{unhinged} (linear), \textit{logistic}, and more. We now introduce our definition of anchor points\footnote{Different notions -related to our definition- of anchor points have been used before in the literature under different names. We review their uses and assumptions in Section \ref{section:related_work}}. 

% What is special about them is that it is a set of instances for which the user provides us with the corresponding \textit{true posterior}, that then can be combined with the equations describing the transformation caused by corruption (Eq.\ref{eq:noisy_posterior}), to provide insights into the value of the noise rates.

\begin{definition}
\emph{(Anchor Points)}
An instance $\boldsymbol{x}$ is called an \textit{anchor point} if we are provided with its true posterior $\eta(\boldsymbol{x})$. Let $\mathcal{A}_{s}^k$ denote a collection of $k$ anchor points, with $\eta(\boldsymbol{x})=s~\forall \boldsymbol{x}\in\mathcal{A}_{s}^k$. Furthermore, let us also define $\mathcal{A}_{s,\delta}^k$, to imply that $\eta(\boldsymbol{x}_i) = s + \epsilon_i$, for $\epsilon_i \sim \mathbb{U}([-\delta,~\delta])$, with $0 \leq \delta \ll 1$ (respecting $0 \leq \eta(\boldsymbol{x}) \leq 1$). Also let $\mathcal{A}_{s,\delta} = \mathcal{A}_{s,\delta}^1$.
\end{definition}

% \begin{wrapfigure}{R}{0.50\textwidth}
%   \vspace{-20pt}
% %   \begin{center}
%   \includegraphics[scale=0.30]{figures/power_and_k.pdf}
% %   \end{center}
%   \vspace{-10pt}
%   \caption{Power of the hypothesis test as a function of $\beta-\alpha$, for a range of different $k$s. We set $v = (\sfrac{1}{16})\cdot(X^\top DX)\inv = 0.1$.}
%   \vspace{-20pt}
% \end{wrapfigure}

\begin{wrapfigure}{r}{8cm}
\vspace{-20pt}
    \begin{empheq}[box=\widefbox]{align*}
        \mathcal{A}_{1}^k~~~&\to~~~\eta(\boldsymbol{x})=1~~~~~\to~~~\tilde{\eta}(\boldsymbol{x})=1-\alpha\\
        \mathcal{A}_{\sfrac{1}{2}}^k~~~&\to~~~\eta(\boldsymbol{x})=\sfrac{1}{2}~~~\to~~~\tilde{\eta}(\boldsymbol{x})=\frac{1-\alpha+\beta}{2}\\
        \mathcal{A}_{0}^k~~~&\to~~~\eta(\boldsymbol{x})=0~~~~~\to~~~\tilde{\eta}(\boldsymbol{x})=\beta
    \end{empheq}
\vspace{-25pt}
\end{wrapfigure}

% \begin{empheq}[box=\othermathbox]{align}
% \sum\limits_{n=1}^{\infty} \frac{1}{n} &= \infty.\\
% \int x^2 ~\text{d}x &= \frac13 x^3 + c.
% \end{empheq} 

The cases we will be referring to are shown to the right.
The first and last, $\mathcal{A}_{1}^k$ and $\mathcal{A}_{0}^k$, have been used in the past in different scenarios. In this work we will make use of the second case, $\mathcal{A}_{\sfrac{1}{2}}^k$. 
% These instances should be understood as \textit{providing no information} with regards to the label.

%% file: sections/tests.tex
\section{Hypothesis Tests based on anchor points}
\label{section:hypothesis_tests}

% \begin{wraptable}{R}{5cm}
% % \begin{table}[!h]
% % \centering
% \begin{tabular}{ccc}
% \cmidrule[1.5pt]{1-3}
%                                  & \textbf{Retain}~$\mathcal{H}_0$   & \textbf{Reject}~$\mathcal{H}_0$                                                  \\ \cmidrule[1.5pt]{1-3}
% \multicolumn{1}{c}{$\mathcal{H}_0$~\textbf{True}}  & Correct       & \begin{tabular}[c]{@{}c@{}}Type I Error\\ (Eq.\ref{eq:sigificance})\end{tabular} \\ \hline
% \multicolumn{1}{c}{$\mathcal{H}_0$~\textbf{False}} & Type II Error & \begin{tabular}[c]{@{}c@{}}Correct\\ (Eq.\ref{eq:power})\end{tabular}      \\ \hline
% \end{tabular}
% \caption{Identifying Type I and Type II errors.}
% \label{table:type_errors}
% % \end{table}
% \end{wraptable}

In this section we introduce our framework for devising hypothesis tests to examine the presence of class-conditional label noise in a given dataset (with uniform noise, as the alternative), assuming we are provided with an anchor point(s). Our procedure is based on a two-sided \textit{z-test} (see for example Chapter 8 of \cite{CaseBerg:01}) with a simple null hypothesis, and a composite alternative hypothesis (Eq.\ref{eq:htest_noise_rates}). We first define the distribution under the null hypothesis (Eq.\ref{eq:htest_null}), and under the alternative hypothesis (Eq.\ref{eq:htest_alternative}), when provided with one strict anchor point ($\eta(x) = \sfrac{1}{2}$). In this setting, for a fixed \textit{level of significance} (Type I error) (Eq.\ref{eq:sigificance}), we first derive a region for retaining the null hypothesis (Eq.\ref{eq:null_retain}), and then we analyse the \textit{power} (Prop.\ref{PROP:POWER}) of the test (where we have that Type II Error = 1 - \textit{power}). We then extend the approach to examine scenarios that include: (1) having multiple strict anchors ($\eta(x_i) = \sfrac{1}{2},~\forall i \in [k],~k>1$), (2) having multiple relaxed anchors ($\eta(x_i) \approx \sfrac{1}{2},~\forall i \in [k],~k>1$), and (3) having no anchors.

With the application of the \textit{delta method} (See for example Chapter 3 of \cite{van2000asymptotic}) on Eq.\ref{eq:mle_asymptotic}, we can get an asymptotic distribution for the predictive posterior:
\begin{equation}
    \label{eq:asymptotic_posterior}
    \sqrt{n}(\hat{\eta}(\boldsymbol{x})-\eta(\boldsymbol{x})) \stackrel{D}{\longrightarrow} \mathcal{N}\left(0,~[\eta(\boldsymbol{x})\{1-\eta(\boldsymbol{x})\}]^{2}\cdot \boldsymbol{x}^{\top} \boldsymbol{I}_{\theta_{0}}\inv \boldsymbol{x}\right)
\end{equation}

This fails in the case of $\eta(\boldsymbol{x}) \in \{0,~1\}$, so instead we work with $\sfrac{1}{2}$. Which, together with the approximation of the Fisher-Information matrix with the empirical Hessian, we get:

\begin{equation}
    \hat{\eta}(\boldsymbol{x}) \stackrel{D}{\longrightarrow} \mathcal{N}\left(\frac{1}{2},~\frac{1}{16}\cdot \boldsymbol{x}^{\top} \hat{H}_n \boldsymbol{x}\right)
\end{equation}
where $\hat{H}_n = (X^\top DX)\inv$, where $D$ is a diagonal matrix, with $D_{ii} = \hat{\eta}_i(1 - \hat{\eta}_i)$, where $\hat{\eta}_i = \sigma(\boldsymbol{x}_i^\top\hat{\theta})$.

For the settings: $(\mathcal{D},~\mathcal{A}_{\sfrac{1}{2}}^k)$ and $(\mathcal{D}_{\tau},~\mathcal{A}_{\sfrac{1}{2}}^k)$, for an $\boldsymbol{x} \in \mathcal{A}_{\sfrac{1}{2}}^k$ we get: $\tilde{\eta}(\boldsymbol{x}) = \frac{1}{2}$. While for $(\mathcal{D}_{\alpha, \beta},~\mathcal{A}_{\sfrac{1}{2}}^k)$ we get: $\tilde{\eta}(\boldsymbol{x}) = \frac{1 - \alpha + \beta}{2}$. Note that under $(\mathcal{D}_{\tau},~\mathcal{A}_{\sfrac{1}{2}}^k)$, we also have $[\tilde{\eta}(\boldsymbol{x})\{1-\tilde{\eta}(\boldsymbol{x})\}]^{2} = \frac{1}{16}$ similarly to $(\mathcal{D},~\mathcal{A}_{\sfrac{1}{2}}^k)$.

% We now have all the necessary ingredients to define our two-sided hypothesis test.

\subsection{A Hypothesis Test for Class-Conditional Label Noise}

We now define our null hypothesis ($\mathcal{H}_0$) and (implicit) alternative hypothesis ($\mathcal{H}_1$) as follows:
\begin{equation}
  \label{eq:htest_noise_rates}
  \mathcal{H}_{0}: \alpha = \beta~~~~~\&~~~~~\mathcal{H}_{1}: \alpha \neq \beta
\end{equation}
% \begin{equation}
% \label{eq:htest_noise_rates}
%   \begin{split}
%     & \mathcal{H}_{0}: \alpha = \beta \\
%     & \mathcal{H}_{1}: \alpha \neq \beta
%   \end{split}
% \end{equation}

Under the null and the alternative hypotheses, we have the following distributions for the estimated posterior of the anchor:
\begin{align}
    \mathcal{H}_{0}: \hat{\eta}(\boldsymbol{x}) &\sim \mathcal{N}\left(\frac{1}{2},~\frac{1}{16}\cdot \boldsymbol{x}^{\top} \hat{H} \boldsymbol{x}\right) = \mathcal{N}\left(\frac{1}{2},~v(\boldsymbol{x})\right)\label{eq:htest_null}\\
    \mathcal{H}_{1}: \hat{\eta}(\boldsymbol{x}) &\sim \mathcal{N}\left(\frac{1-\alpha+\beta}{2},~\frac{[(1-\alpha+\beta)(\beta-\alpha)]^2}{16}\cdot \boldsymbol{x}^{\top} \hat{\tilde{H}} \boldsymbol{x}\right) = \mathcal{N}\left(\frac{1+\alpha-\beta}{2},~\tilde{v}(\boldsymbol{x})\right)\label{eq:htest_alternative}
\end{align}

%We now look at understanding the power of the test.
\paragraph{Level of Significance and Power of the test} The \textit{level of significance} (also known as Type I Error) is defined as follows:
\begin{equation}
    \label{eq:sigificance}
    a = \mathbb{P}(\textrm{reject}~ \mathcal{H}_0~|~\mathcal{H}_0~\textrm{is True})
\end{equation}

% \begin{equation*}
%     \frac{\hat{\eta}(\boldsymbol{x}) - \sfrac{1}{2}}{\sqrt{v(\boldsymbol{x})}} \sim \mathcal{N}(0,~1)\\
% \end{equation*}

Rearranging Eq.\ref{eq:htest_null} we get: $\frac{\hat{\eta}(\boldsymbol{x}) - \sfrac{1}{2}}{\sqrt{v(\boldsymbol{x})}} \sim \mathcal{N}(0,~1)$, under the null. Which for a chosen level of \textit{significance} ($a$) allows us to define a region of retaining the null $\mathcal{H}_0$. We let $z_{\sfrac{a}{2}}$ and $z_{1-\sfrac{a}{2}}$ denote the lower and upper critical values for retaining the null at a level of significance of $a$.

% \paragraph{Retain $\mathcal{H}_0$ if:}
% \begin{align}
%     ~z_{\sfrac{a}{2}}~&\leq~\frac{\hat{\eta}(x) - \sfrac{1}{2}}{\sqrt{v(x)}}~\leq~z_{1-\sfrac{a}{2}}\nonumber\\
%     ~z_{\sfrac{a}{2}}\cdot \sqrt{v(x)} + \sfrac{1}{2}~&\leq ~~~~~~\hat{\eta}(x)~~~~~\leq~z_{1-\sfrac{a}{2}}\cdot\sqrt{v(x)} + \sfrac{1}{2}\label{eq:null_retain}
%     % ~L~~&\leq ~~~~~~~\hat{\eta}(x)~~~~~~\leq~~U\label
% \end{align}

\paragraph{Retain $\mathcal{H}_0$ if:}\begin{equation}
    ~z_{\sfrac{a}{2}}\cdot \sqrt{v(x)} + \sfrac{1}{2}~\leq ~~~~~~\hat{\eta}(x)~~~~~\leq~z_{1-\sfrac{a}{2}}\cdot\sqrt{v(x)} + \sfrac{1}{2}
    \label{eq:null_retain}
\end{equation}
% \begin{align}
%     % ~z_{\sfrac{a}{2}}~&\leq~\frac{\hat{\eta}(x) - \sfrac{1}{2}}{\sqrt{v(x)}}~\leq~z_{1-\sfrac{a}{2}}\nonumber\\
%     ~z_{\sfrac{a}{2}}\cdot \sqrt{v(x)} + \sfrac{1}{2}~&\leq ~~~~~~\hat{\eta}(x)~~~~~\leq~z_{1-\sfrac{a}{2}}\cdot\sqrt{v(x)} + \sfrac{1}{2}\label{eq:null_retain}
%     % ~L~~&\leq ~~~~~~~\hat{\eta}(x)~~~~~~\leq~~U\label
% \end{align}

Using the region of retaining the null hypothesis, we can now derive the \textit{power} of the test.

\begin{proposition}\label{PROP:POWER}
\emph{Power of the test} (See Appendix 8.3 for the full derivation.)\\
Under the distributions for the estimated posterior under the null and alternative hypotheses in Eqs.\ref{eq:htest_null}\&\ref{eq:htest_alternative}, based on the definition of the hypotheses in Eq.\ref{eq:htest_noise_rates}, the test has power: $\mathbb{P}(\textrm{reject}~ \mathcal{H}_0~|~\mathcal{H}_0~\textrm{is False}) = 1 - b_1$, where:
\begin{align}
    % &1 - b_1 = \mathbb{P}(\textrm{reject}~ \mathcal{H}_0~|~\mathcal{H}_0~\textrm{is False})\label{eq:power}\\
    &b_1 = \Phi\left(\ffrac{z\cdot\sqrt{v(x)} + \frac{\beta-\alpha}{2}}{\sqrt{\tilde{v}(x)}}\right) - \Phi\left(\ffrac{-z\cdot\sqrt{v(x)} + \frac{\beta-\alpha}{2}}{\sqrt{\tilde{v}(x)}}\right)\label{eq:one_minus_power}
\end{align}
\end{proposition}

\subsection{Multiple Anchor Points}

\begin{wrapfigure}{R}{0pt}
    \vspace{-20pt}
    \includegraphics[scale=0.30]{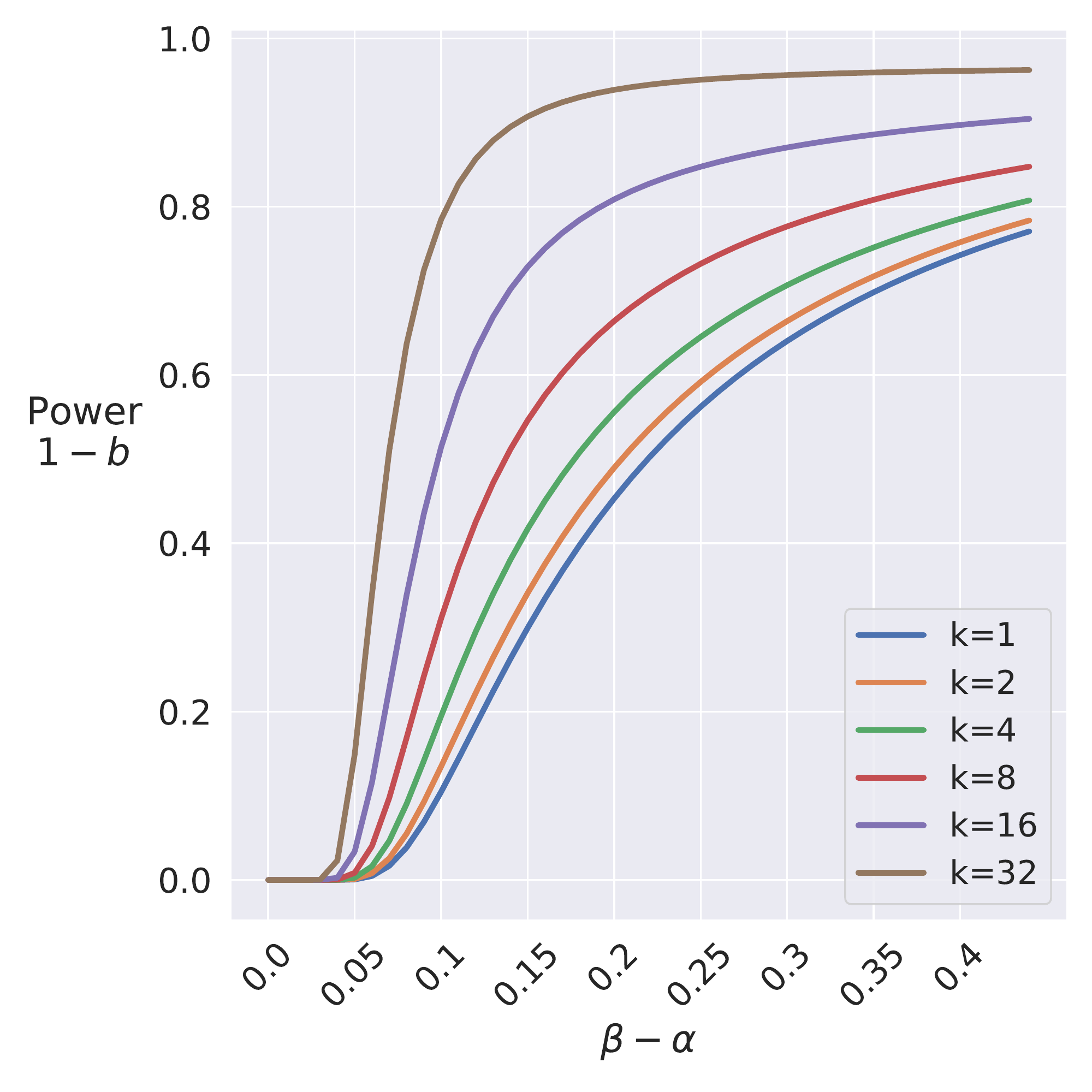}
    \caption{Power of the hypothesis test as a function of $\beta-\alpha$, for a range of different $k$s. We set $v = (\sfrac{1}{16})\cdot(X^\top DX)\inv = 0.1$.}
    \label{fig:power_comparison}
  \vspace{-30pt}
\end{wrapfigure}

In this section we discuss how the properties of the test change in the setting where multiple anchors points are provided. 

% The following expectation ($\mathbb{E}$), and variance ($\mathbb{V}$) operators, unless specified otherwise, are with respect to randomness in $\hat{\theta}_{mle}$ (see Eq.\ref{eq:mle_asymptotic}).

Let $\hat{\eta}_i$ correspond to the $i$th instance in $\mathcal{A}_{\sfrac{1}{2}}^k$. Then for $\bar{\eta} = \frac{1}{k}\sum_{i=1}^k\hat{\eta}_i$ we have:
\begin{equation*}
    \bar{\eta} \sim \mathcal{N}\left(\frac{1}{2},~\frac{1}{16}\cdot \bar{\boldsymbol{x}}^\top H\bar{\boldsymbol{x}}\right)
\end{equation*}
where $\bar{\boldsymbol{x}} = \frac{1}{k}\sum_{i=1}^k \boldsymbol{x}_i$ with $\boldsymbol{x}_i \in \mathcal{A}_{\sfrac{1}{2}}^k~\forall i$. For the full derivation see Appendix 8.4.

\paragraph{Anchors chosen at random} We have that $\boldsymbol{x} \in \mathcal{A}_{\sfrac{1}{2}}^k\to\boldsymbol{x}^\top\beta_0 = 0$, so that for an orthonormal basis $\boldsymbol{U}$, $\boldsymbol{x} = \boldsymbol{U}\boldsymbol{r}$. Without loss of generalisation we let $\boldsymbol{U}_{:,0} = \frac{\beta_0}{\|\beta_0\|_{2}}$, and therefore $\eta(x) = \sfrac{1}{2} \to \boldsymbol{r}_0 = 0$. In words: $\forall \boldsymbol{x} \in \mathcal{A}_{\sfrac{1}{2}}^k$ we have that $\boldsymbol{x}$'s component in the direction of $\beta_0$ is $0$. 

Now we make the assumption that $\boldsymbol{x}$'s are random with $r_j \sim \mathbb{U}([-c,~c])$. Therefore, $\mathbb{E}r_j = 0$, and $\mathbb{V}r_j = \frac{c^2}{3}$. In the following we use the subscript $S$ in the operator $\mathbb{E}_{S}$ to denote the randomness in choosing the set $\mathcal{A}$. In words: we assume that the set $\mathcal{A}_{\sfrac{1}{2}}^k$ is chosen uniformly at random from the set of all anchor points.

Combining these we get:
\begin{align*}
    \mathbb{E}_{S}v(x) &= \mathbb{E}_{S} x^\top Hx = \mathbb{E}_{S} r^\top UHU^\top r\\
    &= \frac{dc^2}{3}\cdot tr(UHU^\top) = \frac{dc^2q}{3}
\end{align*}
% \begin{equation*}
%         \mathbb{E}_{S}v(x) = \mathbb{E}_{S} x^\top Hx = \mathbb{E}_{S} r^\top UHU^\top r
%     = \frac{dc^2}{3}\cdot tr(UHU^\top) = \frac{dc^2q}{3}
% \end{equation*}

where $q=tr(H)$. While for $k$ anchor points chosen independently at random, we get:
\begin{equation*}
    \mathbb{E}_{S}v(\bar{x}) = \mathbb{E}_{S}\left[\frac{1}{k^2}\sum_{i,j}^kx_i^\top Hx_j\right] = \mathbb{E}_{S} \left[\frac{1}{k^2}\sum_{i,j}^k r_i^\top UHU^\top r_j\right]
    = \frac{dc^2}{3k}\cdot tr(UHU^\top) = \frac{dc^2q}{3k}
\end{equation*}

Following the same derivation as above we get:
\begin{equation*}
    b_k = \Phi\left(\ffrac{z\cdot\sqrt{v(\bar{x})} + \frac{\beta-\alpha}{2}}{\sqrt{\tilde{v}(\bar{x})}}\right)~-~\Phi\left(\ffrac{-z\cdot\sqrt{v(\bar{x})} + \frac{\beta-\alpha}{2}}{\sqrt{\tilde{v}(\bar{x})}}\right)
\end{equation*}

If we let $v = \mathbb{E}_Sv(x)$ (similarly $\tilde{v} = \mathbb{E}_S\tilde{v}(x)$), then we have seen that $\mathbb{E}_Sv(\bar{x}) = \frac{v}{k}$ (Reminder: expectations are with respect to the randomness in picking the anchor points). Then we have:
\begin{equation}
    \label{eq:power_comparison}
    \frac{b_k}{b_1} = \ffrac{\Phi\left(\frac{z\sqrt{v} + h\sqrt{k}}{\sqrt{\tilde{v}}}\right)~-~\Phi\left(\frac{-z\sqrt{v} + h \sqrt{k}}{\sqrt{\tilde{v}}}\right)}{\Phi\left(\frac{z\sqrt{v} + h}{\sqrt{\tilde{v}}}\right)~-~\Phi\left(\frac{-z\sqrt{v} + h}{\sqrt{\tilde{v}}}\right)}~~\leq~~1
\end{equation}
with $h = \frac{\beta-\alpha}{2}$.

In Fig.\ref{fig:power_comparison} we compare the power ($1 - b$) of the test, as a function of the difference between the noise rates ($\beta - \alpha$), and number of anchor points used ($k$). We observe that a larger number of anchor points leads to a higher value for power.

\subsection{Multiple Relaxed Anchors-Points}
\label{section:mult_relaxed_anchors}

In this section we see how the properties of the test change in the setting where the anchors do not have a perfect $\eta(\boldsymbol{x}) = \sfrac{1}{2}$. We now consider the case of $\mathcal{A}_{\sfrac{1}{2}, \delta}^k$. Let $\boldsymbol{x}$ be such that $\eta(\boldsymbol{x}) = \frac{1}{2} + \epsilon$, where $\epsilon \sim \mathbb{U}([-\delta, \delta])$ with $0 < \delta \ll 1$. (Note: by definition $\delta \leq \sfrac{1}{2}$.)

For one instance we have the following:~~$\mathbb{E}_{\hat{\theta}}\hat{\eta} = \sfrac{1}{2} + \epsilon,~~~~\text{and}~~~~\mathbb{E}_{S}\mathbb{E}_{\hat{\theta}}\hat{\eta} = \sfrac{1}{2}$
% \begin{align*}
%     \mathbb{E}_{\hat{\theta}}\hat{\eta} = \sfrac{1}{2} + \epsilon,~~~~&\text{and}~~~~\mathbb{E}_{S}\mathbb{E}_{\hat{\theta}}\hat{\eta} = \sfrac{1}{2}
% \end{align*}

For the variance component we have: $[\hat{\eta}(1-\hat{\eta})]^2 = \left[\left(\frac{1}{2} + \epsilon\right)\left(\frac{1}{2} - \epsilon\right)\right]^2  \approx \frac{1}{16} - \frac{\epsilon^2}{2}$,~ignoring terms of order higher than $\epsilon^2$, under the assumption that $\delta \ll 1$. 

Under the \textit{law of total variance} we have:
\begin{align}
    \mathbb{V}(\eta) &= \mathbb{E}\left(\mathbb{V}\left(\eta~|~\epsilon\right)\right) + \mathbb{V}\left(\mathbb{E}\left(\eta~|~\epsilon\right)\right)\nonumber\\
    &=\mathbb{E}\left(\left(\frac{1}{16} - \frac{\epsilon^2}{2}\right)\cdot \boldsymbol{x}^\top H\boldsymbol{x}\right) + \mathbb{V}\left(\frac{1}{k}\sum_{i=1}^k\hat{\eta}_i\right)\nonumber\\
    &=\left(\frac{1}{16} - \frac{\delta^2}{6}\right)\cdot \boldsymbol{x}^\top H\boldsymbol{x}+ \mathbb{V}\left[\frac{1}{2} + \frac{1}{k}\sum_{i=1}^k\epsilon_i\right]\nonumber\\
    &=\left(\frac{1}{16} - \frac{\delta^2}{6}\right)\cdot \boldsymbol{x}^\top H\boldsymbol{x} + \frac{\delta^2}{3k}\label{eq:law_of_total_variance}
\end{align}

For the full derivation see Appendix 8.6. Finally, bringing everything together and ignoring $\delta^2$ terms we get:
\begin{equation*}
    \bar{\eta} \sim \mathcal{N}\left(\frac{1}{2},~\left(\frac{1}{16} - \frac{\delta^2}{6}\right)\cdot \bar{\boldsymbol{x}}^\top H\bar{\boldsymbol{x}}\right) \approx \mathcal{N}\left(\frac{1}{2},~\frac{1}{16}\cdot \bar{\boldsymbol{x}}^\top H\bar{\boldsymbol{x}}\right)
\end{equation*}

\subsection{What if we have no anchor points?}
We have shown that we can relax the hard constraint on the anchor points to be exactly $\eta = \sfrac{1}{2}$, to $\eta \approx \sfrac{1}{2}$. It is natural then to ask if we need anchor points at all. If instead we were to sample points at random, then we would have the following: $\mathbb{E}_{p(X)}\eta(X) = \pi$.
% \begin{equation}
%     \label{eq:no_anchorpoints}
%     \mathbb{E}_{p(X)}\eta(X) = \pi
% \end{equation}
The importance of needing for set of anchor points, either $\mathcal{A}_{\sfrac{1}{2}}^k$ or $\mathcal{A}_{\sfrac{1}{2}, \delta}^k$, is that, the anchor points would be centered around a known value ~$\sfrac{1}{2}$~, as opposed to having no anchor points and sampling at random, where the anchor points would end up being centered around $\pi$. Knowledge of the class priors could allow for a different type of hypothesis tests to asses the presence of label noise. We do not continue this discussion in the main document as it relies on different type of information, but provide pointers in the Appendix 8.8.

\subsection{Practical Considerations \& Limitations}

\vspace{-7pt}
\paragraph{Beyond Logistic Regression}
Our approach relies on the asymptotic properties of MLE estimators, and specifically of Logistic Regression. More complex models can be constructed in a similar fashion through polynomial feature expansion. However the extension of these tests to richer model-classes, such as Gaussian Processes, remains open. 

\vspace{-7pt}
\paragraph{Multi-class classification}
Multi-class classification setting can be reduced to \textit{one-vs-all}, \textit{all-vs-all}, or more general error-correcting output codes setups as described in \cite{10.5555/1622826.1622834}, which rely on multiple runs of binary classification. In these settings then we could apply the proposed framework. The challenge would then be how to interpret $\eta = \sfrac{1}{2}$.

\vspace{-7pt}
\paragraph{Finding anchor points}
While it might not be straightforward for the user to provide instances whose true posterior is $\eta(\boldsymbol{x}) = \sfrac{1}{2}$, we do show how this can be relaxed, by allowing $\eta(\boldsymbol{x}) \approx \sfrac{1}{2}$. We then show how multiple anchor points can be stacked, improving the properties of the test.

\vspace{-7pt}
\paragraph{Model Misspecification}
Our work relies on properties of the MLE and its asymptotic distribution (Eq. \ref{eq:mle_asymptotic}). These assume the model is \textit{exactly} correct. Similarly, under the null in the scenario of $\alpha=\beta>0$, we are at risk of model misspecification. This is not a new problem for Maximum Likelihood estimators, and one remedy is the so-called \textit{Huber Sandwich Estimator} \cite{freedman2006so} which replaces the Fisher Information Matrix, with a more robust alternative.

\vspace{-7pt}
\paragraph{Instance-dependent Noise (IDN)}
% UN is the most basic scenario where learning can be unbiased under mild conditions. The next step is CCN, where learning is in general biased. A motivating example of CCN is given in Frénay and Verleysen [2013] in the form of medical case-control studies, where different tests may be used for subject and control. IDN is a generalisation where the probability of label flipping is allowed to depend on the features. This is the first work to introduce principled quality check on the labels of a dataset and it could serve as a starting point to devise tests of IDN.

In IDN the probability of label flipping depends on the features. It can be seen as a generalisation over UN (which is unbiased under mild conditions (See Theorem \ref{theorem:un_robustness}) and CCN (where learning is in general biased). Our theoretical framework for CCN serves as a starting point to devise tests of IDN. One potential way of extending our approach to test for IDN could be to have anchor points at different contours, i.e., 0.20, 0.30, ... 0.70. 

% A motivating example of CCN is given in \cite{frenay2013classification} in the form of medical case-control studies, where different tests may be used for subject and control.

% A link between CCN and IDN can be seen in the motivating example of Frénay and Verleysen [2013] in the form of medical case-control studies where different tests may be used for different patient groups.

%% file: sections/related_work.tex
\section{Related Work}
\label{section:related_work}
There exist multiple works in the field of weak supervision where instead of being provided with the true labels, the dataset is annotated with a weak version of them, usually derived from the true label and potentially influenced by exogenous variables. It is beyond the scope of this work to discuss this field but these works and the references therein offer an overview of the field: \cite{patrini2016weakly,menon2015learning, cid2014prop, perello2017weak, frenay2013classification}.

We briefly discuss approaches in the literature that relate to tackling the problem of learning with the presence of (flipping) noise in the labels. As already discussed in Theorem \ref{theorem:un_robustness}, in the case of uniform noise, under mild assumptions, we have robust risk minimisation. However, in the case of class-conditional noise, we do not have similar guarantees.

One common approach is to proceed by correcting the loss to be minimised, by introducing the \textit{mixing matrix} $\boldsymbol{M} \in [0, 1]^{c\times c}$, where $M_{i,j} = \mathbb{P}(\tilde{y}=\boldsymbol{e}^j~|~y=\boldsymbol{e}^i)$ \cite{patrini2016weakly}.

\vspace{-7pt}
\paragraph{Anchor points and perfect samples} Using these formulations, we are in a position where, if we have access to $\boldsymbol{M}$, we can correct the training procedure to obtain an unbiased estimator. However, $\boldsymbol{M}$ is rarely known and is difficult to estimate. Works on estimating $\boldsymbol{M}$ rely on having access to `perfect samples' and can be traced back to \cite{scott2013classification}, and the idea was later adapted and generalised in \cite{patrini2016weakly,menon2015learning,liu2015classification,perello2020recycling} to the multi-class setting. Interestingly, in \cite{patrini2017making} authors do not explicitly define these perfect samples, but rather assume they do exist in a large enough (validation) dataset $\boldsymbol{X}'$ - obtaining good experimental results. Similarly, \cite{xia2019anchor} also work by not explicitly requiring anchor points, but rather assuming their existence.

%There also exists a different line of works that aim at estimating $\boldsymbol{M}$ by using a clean sample and Expectation Maximisation \cite{perello2020recycling}.

% For completeness, the procedure they follow is as follows:
% \begin{equation*}
% \begin{array}{l}
% \overline{\boldsymbol{x}}^{i}=\operatorname{argmax}_{\boldsymbol{x} \in X^{\prime}} \hat{p}\left(\tilde{\boldsymbol{y}}=\boldsymbol{e}^{i} \mid \boldsymbol{x}\right) \\
% \hat{M}_{i j}=\hat{p}\left(\tilde{\boldsymbol{y}}=\boldsymbol{e}^{j} \mid \overline{\boldsymbol{x}}^{i}\right)
% \end{array}
% \end{equation*}

% An important observation in \cite{menon2015learning} Section 6.3, relating to the use of Logistic Regression:
% for $\tilde{\eta}(\boldsymbol{x}) = \sigma(\boldsymbol{w}^\top\boldsymbol{x} + b)$, if we set $x = u\boldsymbol{w}$, and let $u \to +\infty$ then $\tilde{\eta}(\boldsymbol{x}) \to 1$, regardless of the noise rates ($\alpha$ and $\beta$). Even though this could be potentially circumvented by defining a radius for which $\|\boldsymbol{x}\|_2^2 \leq R^2~\forall x$, this does not affect the current work, as we are concerned with anchor points that have the property: $\eta(x) = \sfrac{1}{2}$.
% % \todo[inline]{find a few more references for the below.}
% \begin{figure}[ht]
%     \centering
%     \includegraphics[scale=0.21]{figures/noise_transform.pdf}
%     \caption{An illustration of how CCN alters the class distributions of two Gaussian class-conditional distributions.}
%     \label{fig:ccn_effect}
% \end{figure}

\vspace{-7pt}
\paragraph{Noisy examples} An alternative line is followed by \cite{northcutt2019confident, northcutt2017confident}, where the aim is to identify the \textit{specific} examples that have been inflicted with noise. This is a non-trivial task unless certain assumptions can be made about the per-class distributions, and their shape. For example, if we can assume that the supports of the two classes do not overlap (i.e. $\eta(x)(1-\eta(x)) \in \{0, 1\}~\forall x$), then we can identify mislabelled instances using per-class densities. If this is not the case, then it would be difficult to differentiate between a mislabelled instance and an instance for which $\eta(x)(1-\eta(x)) \in (0, 1)$. A different assumption could be uni-modality, which would again provide a prescription for identifying mislabelled instances through density estimation tools.

\vspace{-7pt}
\paragraph{Distilled examples} The authors in \cite{cheng2020learning} go in the opposite direction by trying to identify instances that \textit{have not been corrupted} $\to$ the \textit{distilled examples}. As a first step the authors assume knowledge of an upper-bound\footnote{The paper aims at tackling instance-dependent noise.} (Theorem 2 of \cite{cheng2020learning}) which allows them to define sufficient conditions for identifying whether an instance is \textit{clean}. As a second step they aim at estimating the (local) noise rate based on the neighbourhood of an instance (Theorem 3 of \cite{cheng2020learning}).

\vspace{-7pt}
\paragraph{Informative priors} Bayesian Statistics is often concerned with constructing informative prior distributions that reflect expert knowledge. While it might be challenging eliciting information from experts and modeling it quantitatively; it is often a necessary, and useful, step in low-data settings. 

% Methods that rely on experts quantifying their beliefs include the \textit{conditional means} approach of \cite{bedrick1996new} where a prior distribution is derived from the potential outcomes of given input, and \cite{https://doi.org/10.1111/j.0006-341X.2001.00663.x} where priors beliefs are first expressed in weakly informative ranges (see \cite{gelman2008weakly} for a discussion.) Other works rely on constructing priors based on the specification of quantiles (\cite{garthwaite2005statistical}).

Similarly to the first set of works we introduce and exploit \textit{anchor points}, but not for directly estimating the mixing matrix, but rather to devise hypothesis tests to obtain evidence against the null hypothesis: the dataset having been inflicted with uniform noise, as opposed to class-conditional uniform noise.

%% file: sections/experiments.tex
\section{Experiments}
\label{section:experiments}
In order to illustrate the properties of the tests, for the experiments we consider a synthetic dataset where the per-class distributions are Gaussians, with means $[1,~1]^\top$ and $[-1,~-1]^\top$, with identity as scale. For this setup we know that anchor points should lie on the line $y=-x$, and draw them uniformly at random $x \in [-4,~4]$. We analyse the following parameters of interest: 
\begin{enumerate}
    \item \label{enum:n} $N \in [500,~1000,~2000,~5000]$: the training sample size.
    \item $(\alpha - \beta) \in [-0.05,~0.10,~0.20]$: the difference between the per-class noise rates. \label{enum:ab}
    \item $k \in [1,~2,~4,~8,~16,~32]$: the number of anchor points.\label{enum:k}
    \item $\delta \in [0,~0.05,~0.10]$: how relaxed the anchor points are: $\eta(x) \in [0.50-\delta,~0.50+\delta]$.\label{enum:delta}
\end{enumerate}

For all combinations of $N$ and $(\alpha - \beta)$ we perform $500$ runs. In each run, we generate a clean version of the data $\mathcal{D}$, and then proceed by corrupting it to obtain a separate version: $\mathcal{D}_{\alpha, \beta}$. For both datasets, we fit a Logistic Regression model. We sample both the anchor points and relaxed anchor points. Finally, we then compute the z-scores, and subsequently the corresponding p-values\footnote{What we have so far presented is aligned with the Neyman-Pearson theory of hypothesis testing. We have shown how to utilise anchor points to obtain the p-value -- a continuous measure of evidence against the null hypothesis- and then leverage the implicit \textit{alternative hypothesis} of class-conditional noise and a \textit{significance level} to analyse the \textit{power} of the test. In this case, the p-value is the basis of formal decision-making process of rejecting, or failing to reject, the null hypothesis. Differently, in Fisher's theory of significance testing, the p-value is the end-product \cite{perezgonzalez2015fisher}. Both the p-value and the output of the test can be used as part of a broader decision process that considers other important factors.}.

The box-plots should be read as follows: $Q1,~Q2~\&~Q3$ separate the data into $4$ equal parts. The inner box starts (at the bottom) at $Q1$ and ends (at the top) at $Q3$, with the horizontal line inside denoting the median ($Q2$). The whiskers extend to show $Q1 - 1.5\cdot IQR$, and $Q3 + 1.5\cdot IQR$. $IQR$ denotes the \textit{Interquartile Range} and $IQR = Q3 - Q1$. 

% \todo{dashed horozontal lines?}

\begin{figure}
    \centering
    \begin{minipage}{6cm}
        \includegraphics[width=\linewidth]{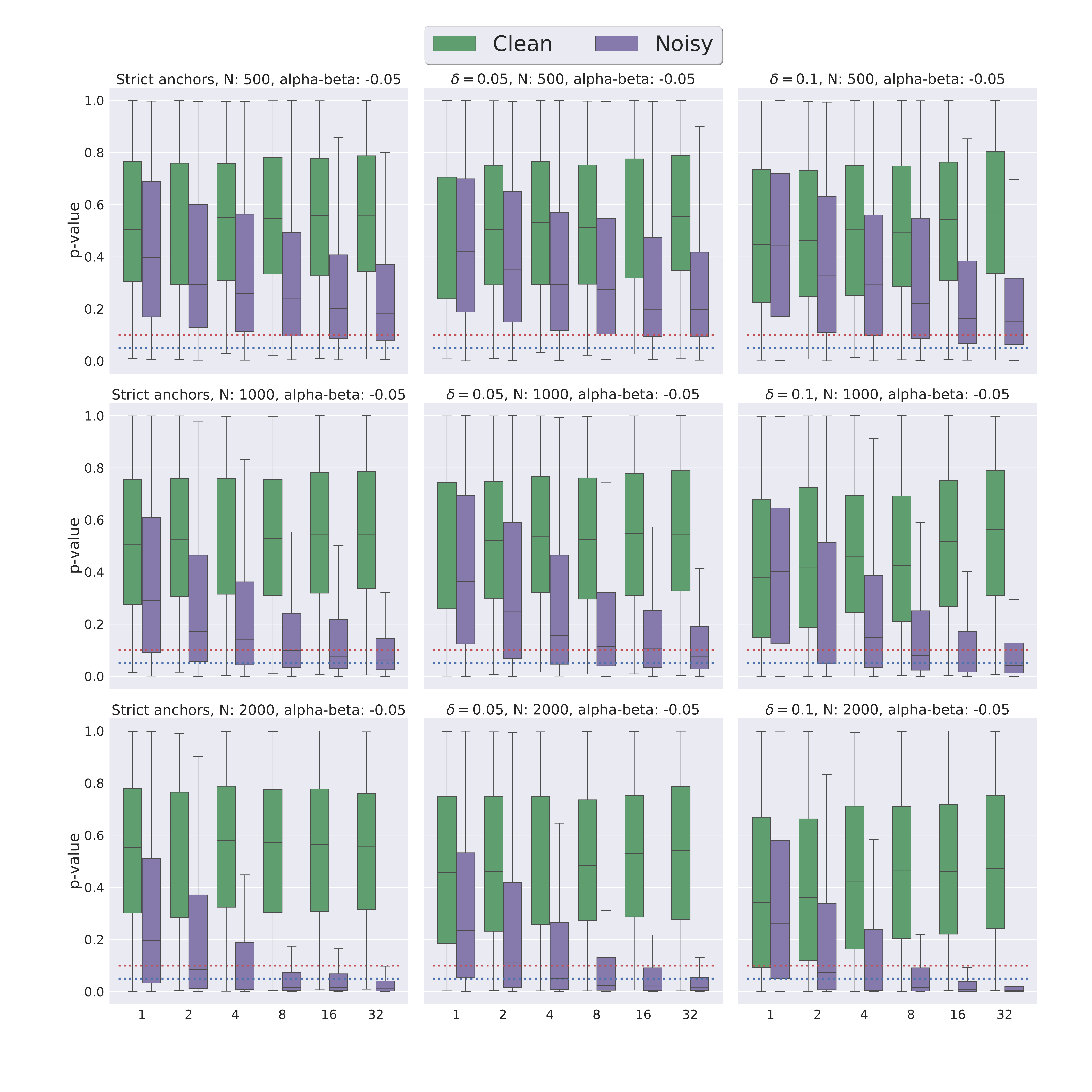}
        \caption{Box-plots with fixed $|\beta - \alpha| = 0.05$. Red dotted line indicates the mark of $0.10$, and the blue one $0.05$.}
        \label{fig:experiments_005}
    \end{minipage}
    \qquad
    \begin{minipage}{6cm}
        \includegraphics[width=\linewidth]{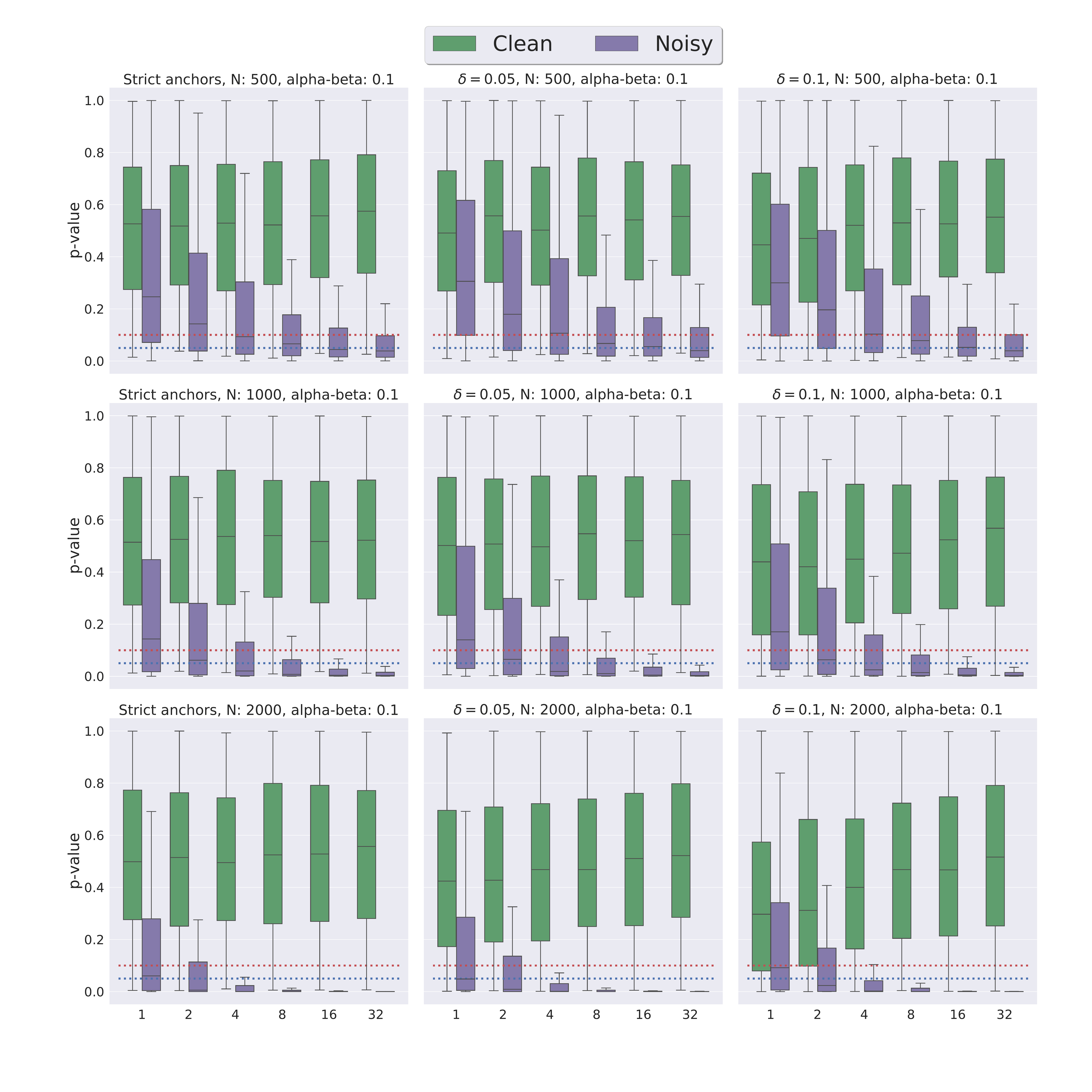}
        \caption{Box-plots with fixed $|\beta - \alpha| = 0.10$. Red dotted line indicates the mark of $0.10$, and the blue one $0.05$.}
        \label{fig:experiments_010}
    \end{minipage}
\end{figure}

\begin{wrapfigure}{R}{0.5\textwidth}
  \includegraphics[width=\linewidth]{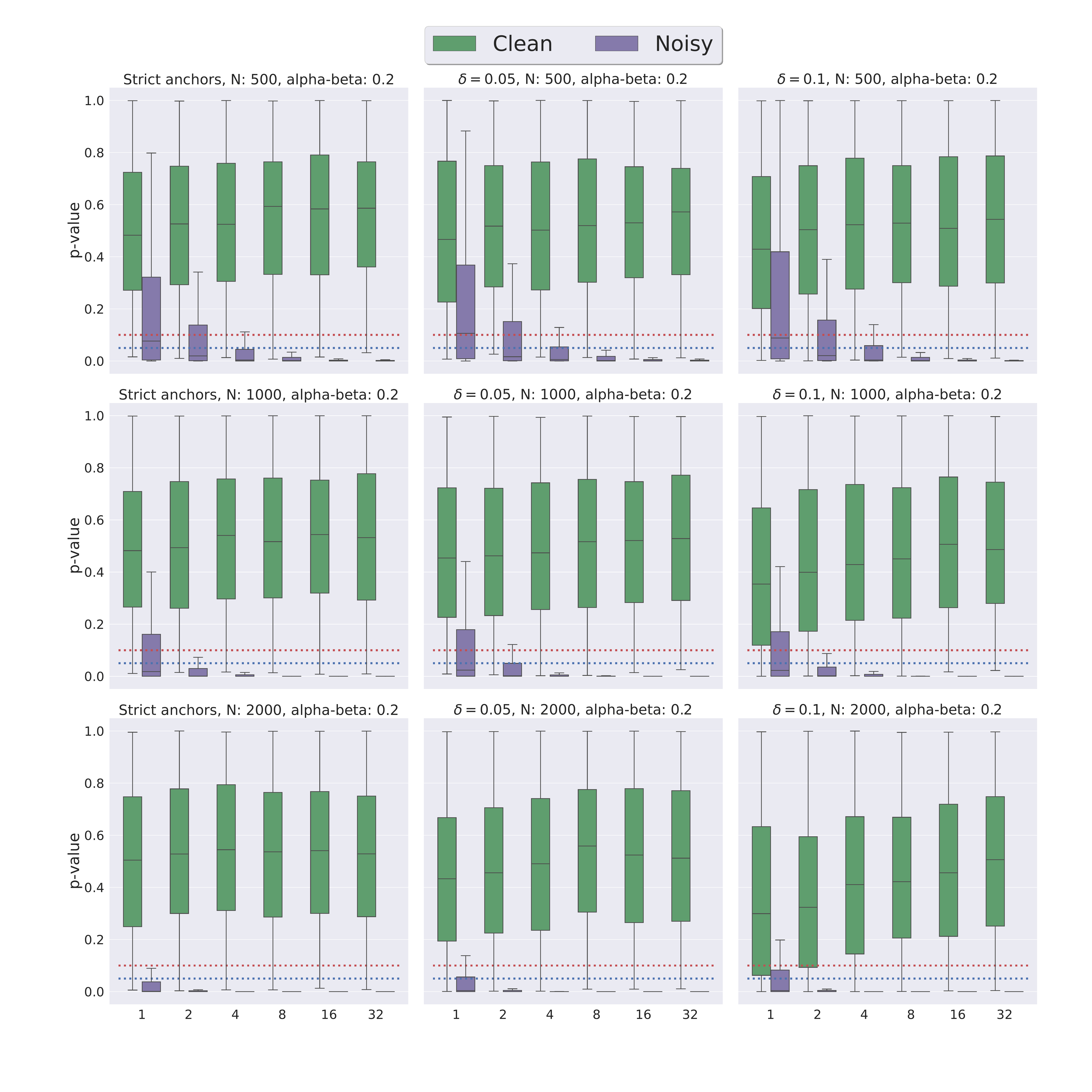}
  \caption{Box-plots with fixed $|\beta - \alpha| = 0.20$. Red dotted line indicates the mark of $0.10$, and the blue one $0.05$.}
  \label{fig:experiments_020}
\end{wrapfigure}

In Figures \ref{fig:experiments_005}, \ref{fig:experiments_010} and \ref{fig:experiments_020} we have the following: moving to the right we increase the relaxation of anchor points, and moving downwards we increase the training sample-size. On the subplot level, on the x-axis we vary the number of anchor points, and on the y-axis we have the p-values. In all subplots we indicate with a red dashed line the mark of $0.10$, and with a blue one the mark of $0.05$, which would serve as rejection thresholds for the null hypothesis.

The experiments are illustrative of the claims made earlier in the paper. Below we discuss the findings in the experiments and what they mean with regards to Type I and Type II errors. We discuss these points in two parts; we first discuss the effect on sample size ($N$), difference in noise rates ($\vert\alpha-\beta\vert$) and number of anchor points ($k$).

\paragraph{Size of training set ($N$)}

As the size of training set ($N$) increases, the power increases. This can be seen Figures \ref{fig:experiments_005}, \ref{fig:experiments_010} \& \ref{fig:experiments_020}. By moving down the first column, and fixing a value for $k$, where $N$ increases, we see the range of the purple box-plots decreasing, and essentially a larger volume of tests falling under the cut-off levels of significance (red and blue dashed lines). This is expected given that the variance of the MLE $\hat{\theta}_{MLE}$ vanishes as $N$ increases, as is seen in Eq.\ref{eq:mle_asymptotic} and discussion underneath it.

\vspace{-7pt}
\paragraph{Difference in noise rates ($\vert\alpha-\beta\vert$)}

As $\vert\alpha-\beta\vert$ increases, the power increases. This can be seen in Figures \ref{fig:experiments_005}, \ref{fig:experiments_010} \& \ref{fig:experiments_020}, by fixing a particular subplot in the first column (for example, top-left one), and a value for $k$, we see again that the volume moves down. As presented in Eq.\ref{eq:one_minus_power}, as $\beta-\alpha$ increases, the power also increases. 

\vspace{-7pt}
\paragraph{Number of anchor points ($k$)}

The same applies to the number of anchor points -- as the number of anchor points ($k$) increases, the power of the test increases. This can be seen in all three figures by focusing in any subplot in the first column, and considering the purple box-plots moving to the right. In Eq.\ref{eq:power_comparison} we see effect of $k$ on the power.\\

In all three discussions above we focused on the first column of each of the figures -- which shows results from experiments on strict anchors. What we also observe in this case (the first column of all figures) is that the p-values follow the uniform distribution under the null (as expected, given the null hypothesis is true) -- shown by the green box-plots. Therefore the portion of Type I Errors = $a$ (the level of significance Eq.\ref{eq:sigificance}). When we relax the requirements for strict anchors to allow for values close to $\sfrac{1}{2}$, we introduce a bias in the lower and upper bounds in Eq.\ref{eq:null_retain} of $+\epsilon$. While $\mathbb{E}\epsilon = 0$ this shift on the boundaries of the retention region will increase Type I Error. On the other hand, in Eq.\ref{eq:law_of_total_variance} we see how this bias decreases as you increase the number of anchor points. Both of these phenomena are also shown experimentally by looking at the latter two columns of the figures.

\vspace{-7pt}
\paragraph{Anchor point relaxation ($\delta$)}

Lastly, we examine the effect of relaxing the strictness of the anchors ($\delta$), $\eta(x) \in [0.50-\delta,~0.50+\delta]$ on the properties of the test. As just discussed we see that as we increase the number of anchor points Type I Error decreases (volume of green box-plots under each of the cut-off points). We also observe that, as compared to only allowing strict anchors, the power is not affected significantly -- with the effect decreasing as the number of anchor points increases. Furthermore, in the latter two columns we also observe the phenomena mentioned in the discussion concerning the first column only.

% \begin{figure}[H]
%     \centering
%     \includegraphics[width=\linewidth]{results/further_reduced_alpha_0.05_beta_0.1.pdf}
%     \caption{Box-plots with fixed $|\beta - \alpha| = 0.05$. Red dotted line indicates the mark of $0.10$, and the blue one $0.05$.}
%     \label{fig:experiments_005}
% \end{figure}

% \begin{figure}[t]
%     % \centering
%     %\includegraphics[scale=0.12]{results/reduced_alpha_0.2_beta_0.1.pdf}
%     \includegraphics[width=\linewidth]{results/further_reduced_alpha_0.2_beta_0.1.pdf}
%     \caption{Box-plots with fixed $|\beta - \alpha| = 0.10$. Red dotted line indicates the mark of $0.10$, and the blue one $0.05$.}
%     \label{fig:experiments_010}
% \end{figure}

% \begin{figure}[!t]
%     \centering
%     \includegraphics[width=\linewidth]{results/further_reduced_alpha_0.3_beta_0.1.pdf}
%     \caption{Box-plots with fixed $|\beta - \alpha| = 0.20$. Red dotted line indicates the mark of $0.10$, and the blue one $0.05$.}
%     \label{fig:experiments_020}
% \end{figure}

% \begin{figure}
%     \centering
%     \includegraphics[width=1.4\textwidth]{results/alpha_0.3_beta_0.1.pdf}
%     \caption{Figure with centering}
% \end{figure}

% \begin{figure}
%     \centering
%     \includegraphics[width=1.4\textwidth]{results/alpha_0.4_beta_0.1.pdf}
%     \caption{Figure with centering}
% \end{figure}

%% file: sections/conclusion.tex
\section{Conclusion \& Future Work}

In this work we introduce the first statistical hypothesis test for class-conditional label noise. Our approach requires the specification of anchor points, i.e. instances whose labels are highly uncertain under the true posterior probability distribution, and we show that the test's significance and power is preserved over several relaxations on the requirements for these anchor points. 

Our experimental analysis, which confirms the soundness of our test, explores many configurations of practical interest for practitioners using this test. Of particular importance for practitioners, since anchor specification is under their control, is the high correspondence shown theoretically and experimentally between the number of anchors and test significance.

% In this work we take a first step into defining statistical hypothesis testing procedures that can be employed to assess the likelihood of the dataset to be inflicted with class-conditional noise. These tests depend on the user (the expert) providing point(s) with a known true posterior of $\eta(x) = \sfrac{1}{2}$, or with a set of points with $\eta(x) \approx \sfrac{1}{2}$. We show experimentally the soundness of the procedure. We also briefly mention the use of a different type of user-provided information, the true class-priors.

Future work will cover both theoretical and experimental components. On the theoretical front, we are interested in understanding the test's value under a richer set of classification models, and further relaxing requirements on true posterior uncertainty for anchor points. Experimentally, we are particularly interested in applying the tests to diagnostically challenging healthcare problems and  utilising clinical experts for anchor specification.

% extending this work to the multi-class setting (and understanding the implications of our requirements in this case), 

% Specifically we will apply the test to help understand and model the trajectory of dementia which is known to be difficult and subjective task. 

%% file: sections/appendix.tex
\section{Appendix}
% \appendix
% Code available at:\\ \verb|https://drive.google.com/file/d/1_mKV9JO215oblRFc2D9U8sIzLEoHvdkA/view?usp=sharing|

\subsection{Derivation of the Noisy Posterior}
\label{derivation:noisy_posterior}
We can compute the noisy posterior: $\tilde{\eta}(x) = \mathbb{P}(\tilde{y} = 1~|~X=x)$ as follows:
\begin{align}
    \tilde{\eta}(x)~&=~\mathbb{P}(\tilde{y}=1,~y=1~|~X=x)\nonumber\\
    &~~~~~~~~+~\mathbb{P}(\tilde{y}=1,~y=0~|~X=x) \nonumber\\
    &=~\mathbb{P}(\tilde{y}=1~|~y=1)\mathbb{P}(y=1~|~X=x)\nonumber\\
    &~~~~~~~~+~\mathbb{P}(\tilde{y}=1~|~y=0)\mathbb{P}(y=0~|~X=x) \nonumber\\
    &=~(1-\alpha)\cdot\eta(x) + \beta\cdot(1-\eta(x)) \nonumber\\
    &=~\left\{\begin{array}{ll}
(1-\alpha-\beta)\cdot\eta(\boldsymbol{x}) + \beta & \text { if } \textrm{(CCN)} \\
(1-2\tau)\cdot\eta(\boldsymbol{x}) + \tau & \text { if } \textrm{(UN)} 
\end{array}\right.\nonumber
\end{align}

\subsection{Sufficient conditions for robustness to uniform noise}
\label{proof:sufficient_conditions}
In the case of uniform-noise we have that:
\begin{align*}
    sign(\tilde{\eta}(x) - 0.50)~&=~sign((1-2\tau)\cdot(\eta(x) - 0.50))\\
    ~&=~sign(\eta(x) - 0.50) 
\end{align*}
where the last equality holds under the assumptions in Def.~\ref{def:uniform_noise}.
\begin{align*}
    &\mathbb{E}_{X, Y}l(Y, f(X)) = \mathbb{E}_{X} \mathbb{E}_{Y | X} l(Y, f(X)) = \mathbb{E}_{X} \mathbb{E}_{Y | X} \psi(Yf(X))\\
    & \text{but we have access to noisy versions of labels} \\
    &\mathbb{E}_{X} \mathbb{E}_{\tilde{Y} | X} \psi(\tilde{Y}f(X)) \\
    &=\mathbb{E}_{X} \Big[ \tilde{\eta} \cdot \psi(f(X)) + (1 - \tilde{\eta}) \cdot \psi(-f(X)) \Big]\\
    &=\mathbb{E}_{X} \Big[ [(1 - \tau)\eta + \tau(1 - \eta)] \cdot \psi(f(X))\\
    &~~~~~+ [(1 - \tau)(1 - \eta) + \tau\eta] \cdot \psi(-f(X)) \Big]\\
    &=\mathbb{E}_{X} \Big[(1-2\tau) \cdot \mathbb{E}_{Y | X} \psi(Yf(X))\\
    &~~~~~+ \tau \cdot [\psi(f(X)) + \psi(-f(X))] \Big]\\
    &=(1 - 2\tau) \cdot \mathbb{E}_{X, Y}l(Y, f(X)) + \tau \cdot BIAS_{\tau}(\psi)
\end{align*}

This implies that if we train under uniform-noise with rate: $\tau < 0.50$, with a loss with the property that $\psi(f(X)) + \psi(-f(X)) = K$ for a constant $K$ then risk minimisation is tolerant to noise \cite{ghosh2015making}.

\subsection{Derivation for proposition~\ref{PROP:POWER}}
\label{proof:power_one_anchor}
We let $L$ and $U$ denote the lower and upper bounds in Eq.\ref{eq:null_retain} respectively, and let $\epsilon \sim \mathcal{N}(0,~1)$.
\begin{align*}
    b_1&~=\mathbb{P}(\textrm{retain}~ \mathcal{H}_0~|~\mathcal{H}_0~\textrm{is False})\\
    &~=\mathbb{P}\left(L \leq \hat{\eta}(x) \leq U~|~\hat{\eta}(x) \sim \mathcal{N}\left(\frac{1+\alpha-\beta}{2},~\tilde{v}(x)\right)\right)\\
    &~=\mathbb{P}\left(L \leq \hat{\eta}(x) \leq U~|~\hat{\eta}(x) = \frac{1+\alpha-\beta}{2} + \sqrt{\tilde{v}(x)}\cdot\epsilon\right)\\
    &~=\mathbb{P}\left(\ffrac{L - \frac{1+\alpha-\beta}{2}}{\sqrt{\tilde{v}(x)}}~\leq~\epsilon~\leq~\ffrac{U - \frac{1+\alpha-\beta}{2}}{\sqrt{\tilde{v}(x)}}\right)\\
    &~=\mathbb{P}\left(\ffrac{-z\cdot\sqrt{v(x)} + h}{\sqrt{\tilde{v}(x)}}~\leq~\epsilon~\leq~\ffrac{z\cdot\sqrt{v(x)} + h}{\sqrt{\tilde{v}(x)}}\right)\\
    &~=\Phi\left(\frac{z\cdot\sqrt{v(x)} + h}{\sqrt{\tilde{v}(x)}}\right) - \Phi\left(\frac{-z\cdot\sqrt{v(x)} + h}{\sqrt{\tilde{v}(x)}}\right)
\end{align*}
where we have used: $h = \frac{\beta-\alpha}{2}$, for ease of notation.

\subsection{Mean \& Variance for multiple anchors-points}
\label{derivation:moments_multiple_anchors}
For the expectation we have:
\begin{equation*}
     \mathbb{E}\left[\frac{1}{k}\sum_{i=1}^k\hat{\eta}_i\right] = \frac{1}{k}\sum_{i=1}^k\mathbb{E}\hat{\eta}_i = \frac{1}{2}
\end{equation*}

And, for the variance we have:
\begin{equation*}
    \mathbb{V}\left[\frac{1}{k}\sum_{i=1}^k\hat{\eta}_i\right] = \frac{1}{k^2}\left[\sum_{i=1}^k \mathbb{V}\hat{\eta}_i + 2\cdot\sum_{i, j>i} Cov(\hat{\eta}_i, \hat{\eta}_j) \right]
\end{equation*}
with
\begin{align*}
    \mathbb{V}\hat{\eta}_i~&=~\frac{1}{16}\cdot \boldsymbol{x}_i^\top H\boldsymbol{x}_i\\
    \textrm{and}~~Cov(\hat{\eta}_i, \hat{\eta}_j)~&=~\frac{1}{16}\cdot x_i^\top Hx_j
\end{align*}
For the derivation of $Cov(\hat{\eta}_i, \hat{\eta}_j)$ see Appendix \ref{appendix:covariance}.

\subsection{Covariance of estimated posteriors for the case of multiple anchor-points}
\label{appendix:covariance}.

In this section we estimate $Cov(\hat{\eta}(x_i), \hat{\eta}(x_j))$ for the multiple anchors.

We will make use of the following: Let $\hat{\theta}$ be such that $\mathbb{E}\hat{\theta}=\theta_0$, then we have:
\begin{align*}
    f(\hat{\theta}) &\approx f(\theta_0) + (\hat{\theta} - \theta_0)^T\nabla f|_{\theta_0} \\
    &~~~~~~~~+ \frac{1}{2}(\hat{\theta} - \theta_0)^T\nabla^2 f|_{\theta_0}(\hat{\theta}-\theta_0)\\
    &= f(\theta_0) + \frac{1}{2}(\hat{\theta} - \theta_0)^T\nabla^2 f|_{\theta_0}(\hat{\theta}-\theta_0)
\end{align*}
Let $\hat{\eta}(x_i) = \hat{\eta}_i$, then we have:
\begin{align*}
    Cov(\hat{\eta}_i, \hat{\eta}_j) &= \mathbb{E}\left[(\hat{\eta}_i - \mathbb{E}\hat{\eta}_i)(\hat{\eta}_j - \mathbb{E}\hat{\eta}_j)\right]\\
    &=\mathbb{E}[\hat{\eta}_i\hat{\eta}_j] - \eta_i\eta_j
\end{align*}

A few useful derivations (the hat ($\hat{\eta}(x)$) is implied), we also let $\nabla = \nabla_{\theta}$:
\begin{enumerate}
    \item $\nabla{\eta_i} = x_i \eta_i(1-\eta_i) = x_i\gamma_i$
    \item $\nabla{\gamma_i} = \nabla[\eta_i - \eta_i^2] = (1-2\eta_i)x_i\gamma_i$
    \item $\nabla[\eta_i\eta_j] = \eta_j\cdot x_i\gamma_i +  \eta_i\cdot x_j\gamma_j$
    \item 
        $\begin{aligned}[t]
            \nabla[\eta_j\cdot x_i\gamma_i] &= x_i\left[\eta_j\nabla\gamma_i + \gamma_i\nabla \eta_j \right]\\
                &= x_i\left[\eta_j\cdot(1-2\eta_i)x_i\gamma_i + \gamma_i\cdot x_j\gamma_j \right]\\
    &= x_ix_i^\top\cdot\eta_j(1-2\eta_i)\gamma_i + x_ix_j^\top\cdot\gamma_i\gamma_j\\
        \end{aligned}$
    \item \label{enum:item_last}
        $\begin{aligned}[t]
            \nabla^2[\eta_i\eta_j] &= x_ix_i^\top\cdot\eta_j(1-2\eta_i)\gamma_i\\
            &~~~+ 2\cdot x_ix_j^\top\cdot\gamma_i\gamma_j\\
            &~~~+ x_jx_j^\top\cdot\eta_i(1-2\eta_j)\gamma_j\\
        \end{aligned}$
\end{enumerate}

by plugging $\eta_i=\eta_j=\sfrac{1}{2}$ in item \ref{enum:item_last} above, combined with Eq.\ref{eq:mle_asymptotic}, and approximation of the Fisher-Information matrix with the hessian, we get:
\begin{equation*}
    Cov(\hat{\eta}(x_i), \hat{\eta}(x_j)) = \frac{1}{16}x_i^\top \hat{H} x_j
\end{equation*}

\subsection{Variance for Multiple Relaxed Anchor-points}
\label{derivation:variance_relaxed_anchors}

For the variance we have:
\begin{equation*}
    \mathbb{V}_{\hat{\theta}}\left[\frac{1}{k}\sum_{i=1}^k\hat{\eta}_i\right] = \frac{1}{k^2}\left[\sum_{i=1}^k \mathbb{V}_{\hat{\theta}}\hat{\eta}_i + 2\cdot\sum_{i, j>i} Cov_{\hat{\theta}}(\hat{\eta}_i, \hat{\eta}_j) \right]
\end{equation*}
with
\begin{align*}
    \mathbb{E}_{S}\mathbb{V}_{\hat{\theta}}\hat{\eta}_i~&=~\left(\frac{1}{16} - \frac{\delta^2}{6}\right)\cdot \boldsymbol{x}_i^\top H\boldsymbol{x}_i\\
    \textrm{and}~~\mathbb{E}_{S}Cov_{\hat{\theta}}(\hat{\eta}_i, \hat{\eta}_j)~&=~\left(\frac{1}{16} - \frac{\delta^2}{6}\right)\cdot x_i^\top Hx_j
\end{align*}
For the derivation of $Cov(\hat{\eta}_i, \hat{\eta}_j)$ see Appendix \ref{appendix:covariance_relaxed}.

\subsection{Covariance of estimated posteriors for the case of multiple relaxed anchor-points}
\label{appendix:covariance_relaxed}

In the section we estimate: $Cov(\hat{\eta}(x_i), \hat{\eta}(x_j))$ for the case of having multiple relaxed anchors.

We continue from Item \ref{enum:item_last} of Appendix \ref{appendix:covariance}:
\begin{align*}
    \nabla^2[\eta_i\eta_j] &= x_ix_i^\top\cdot\eta_j(1-2\eta_i)\gamma_i\\
    &~~~+ 2\cdot x_ix_j^\top\cdot\gamma_i\gamma_j\\
    &~~~+ x_jx_j^\top\cdot\eta_i(1-2\eta_j)\gamma_j
\end{align*}
We let $\eta_i = \sfrac{1}{2} + \epsilon_i$ and $\eta_j = \sfrac{1}{2} + \epsilon_j$, then we have: (for the first and third terms above)
\begin{align*}
    \eta_j(1-2\eta_i)\gamma_i &= \left(\frac{1}{2} + \epsilon_j\right)\left(-2\epsilon_i\right)\left(\frac{1}{2} + \epsilon_i\right)\left(\frac{1}{2} - \epsilon_i\right)\\
    \mathbb{E}_S[\eta_j(1-2\eta_i)\gamma_i] &= \mathbb{E}_S\left[\left(\frac{1}{2} + \epsilon_j\right)(-2\epsilon_i)\left(\frac{1}{2} + \epsilon_i\right)\left(\frac{1}{2} - \epsilon_i\right)\right]\\
    &= \mathbb{E}_S\left[\left(-\epsilon_i\right)\left(\frac{1}{2} + \epsilon_i\right)\left(\frac{1}{2} - \epsilon_i\right)\right]\\
    &= \mathbb{E}_S\left[-\epsilon_i\left(\frac{1}{4} - \epsilon_i^2\right)\right] = 0\\
\end{align*}

For the second term we have:
\begin{align*}
    \gamma_i &= \left(\frac{1}{2} - \epsilon_i\right)\left(\frac{1}{2} + \epsilon_i\right) = \frac{1}{4} - \epsilon_i^2\\
    \mathbb{E}_S\gamma_i\gamma_j &= \mathbb{E}_S\left(\frac{1}{4} - \epsilon_i^2\right)\left(\frac{1}{4} - \epsilon_j^2\right)\\
    &= \frac{1}{16} - \frac{\delta^2}{6} + \delta^4\\
    &\approx \frac{1}{16} -\frac{\delta^2}{6}
\end{align*}

\subsection{Test based on priors}
\label{section:htest_priors}

Another important relationship is that between the clean and noisy class priors: $\tilde{\pi}~=~\mathbb{P}(\tilde{y}~=~1)$:
\begin{align*}
    \mathbb{P}(\tilde{y}~=~1)~&=~\mathbb{P}(\tilde{y}~=~1,~y=1)~+~\mathbb{P}(\tilde{y}~=~1,~y=0) \\
    &=~\mathbb{P}(\tilde{y}~=~1~|~y=1)\mathbb{P}(y=1)~\\
    &~~~~+~\mathbb{P}(\tilde{y}~=~1~|~y=0)\mathbb{P}(y=0) \\
    &=~(1-\alpha)\cdot\pi + \beta\cdot(1-\pi)
\end{align*}
which under the two settings, $UN$ and $CCN$, gives:
\begin{equation}
    \label{eq:noisy_prior}
    \tilde{\pi}~=\left\{\begin{array}{ll}
(1-\alpha-\beta)\cdot\pi + \beta & \text { if } \textrm{(CCN)} \\
(1-2\tau)\cdot\pi + \tau & \text { if } \textrm{(UN)} \\
\end{array}\right.
\end{equation}
The relationships in Eq.\ref{eq:noisy_prior}, combined with the knowledge of the true class priors, would allow someone to carry out Binomial Hypothesis Tests for presence of label noise. These tests would not need to rely on MLE asymptotics.

% \end{appendix}